\pdfoutput=1

\documentclass[11pt,table,xcdraw]{article}

\usepackage[final]{acl}

\usepackage{times}
\usepackage{latexsym}
\usepackage{graphicx}
\usepackage{caption}
\usepackage{subcaption}
\usepackage{multirow}%
\usepackage{amsmath,amssymb,amsfonts}%
\usepackage{amsthm}%
\usepackage[title]{appendix}%
\usepackage{multirow}
\usepackage{multicol}
\usepackage{textcomp}%
\usepackage{manyfoot}%
\usepackage{booktabs}%
\usepackage{algorithm}%
\usepackage{algorithmicx}%
\usepackage{algpseudocode}%
\usepackage{listings}%
\usepackage{soul}
\usepackage{marginnote}
\usepackage[table,xcdraw]{xcolor}
\usepackage{fancyhdr}
\usepackage{fancybox} 
\usepackage[normalem]{ulem}
\newcommand{\hlc}[2][yellow]{{%
    \definecolor{foo}{HTML}{#1}
    \sethlcolor{foo}\hl{#2}}%
}
\useunder{\uline}{\ul}{}

\usepackage[utf8]{inputenc}
\usepackage{fancyhdr}

\usepackage{microtype}

\usepackage{inconsolata}

\usepackage{graphicx}
\usepackage{subcaption}
\title{Beyond Checkmate:\\Exploring the Creative Choke Points for AI Generated Texts}

\author{
Nafis Irtiza Tripto\textsuperscript{1} \hspace{.1cm} 
Saranya Venkatraman\textsuperscript{2}\thanks{Work completed while PhD student at PennState} \hspace{.1cm} 
Mahjabin Nahar\textsuperscript{1} \hspace{.1cm} 
Dongwon Lee\textsuperscript{1} \\
\textsuperscript{1} The Pennsylvania State University \hspace{.1cm} 
\textsuperscript{2} Amazon \hspace{.1cm}  \\ 
\textsuperscript{1}\texttt{\{nit5154, mfn5333, dongwon\}@psu.edu} \hspace{.1cm} \textsuperscript{2} \texttt{saranvn@amazon.com} \\
}

\begin{document}
\fancypagestyle{firstpage}{
  \fancyhf{}  
  \cfoot{To be appeared at the \textit{30th Conference on Empirical Methods in Natural Language Processing} (\textbf{EMNLP'25} Main) Conference, November 4-9, 2025, Suzhou, China}  
}

\thispagestyle{firstpage}  

\maketitle
\begin{abstract}

The rapid advancement of Large Language Models (LLMs) has revolutionized text generation but also raised concerns about potential misuse, making detecting LLM-generated text (AI text) increasingly essential. While prior work has focused on identifying AI text and effectively \textit{checkmating} it, our study investigates a less-explored territory: portraying the nuanced distinctions between human and AI texts across text segments (introduction, body, and conclusion). Whether LLMs excel or falter in incorporating linguistic ingenuity across text segments, the results will critically inform their viability and boundaries as effective creative assistants to humans. Through an analogy with the structure of chess games, comprising opening, middle, and end games, we analyze segment-specific patterns to reveal where the most striking differences lie. Although AI texts closely resemble human writing in the body segment due to its length, deeper analysis shows a higher divergence in features dependent on the continuous flow of language, making it the most informative segment for detection. Additionally, human texts exhibit greater stylistic variation across segments, offering a new lens for distinguishing them from AI. Overall, our findings provide fresh insights into human-AI text differences and pave the way for more effective and interpretable detection strategies. Codes available at \url{https://github.com/tripto03/chess_inspired_human_ai_text_distinction}.
\end{abstract}

\section{Introduction}

When Garry Kasparov, then world chess champion, lost to IBM’s Deep Blue, a chess-playing supercomputer, in 1997 \cite{pandolfini1997kasparov}, it marked a turning point in AI history, the moment machines overtook humans in a game long considered a symbol of strategic mastery. A similar shift occurred with the public debut of ChatGPT in late 2022, as Large Language Models (LLMs) captured global attention and began reshaping the landscape of communication, creativity, and cognition. With models like \textit{GPT-4} passing professional exams \cite{katz2024gpt_bar_exam} and even approaching Turing test benchmarks \cite{jones2025turing_test}, these advancements raise critical questions about distinctiveness of human intellect. Interestingly, AI chess engines and LLMs share a remarkable similarity. While chess engines determine the best move from a given board state, LLMs predict the next token based on preceding text. This shared mechanism of context-driven prediction has even led to the development of transformer-based chess engines capable of achieving Grandmaster-level performance \cite{ruoss2024grandmaster} and evaluating LLM's capability in Chess \citep{guo2024can, wang2025explore}.

\begin{figure}
    \centering
    \includegraphics[width=\linewidth]{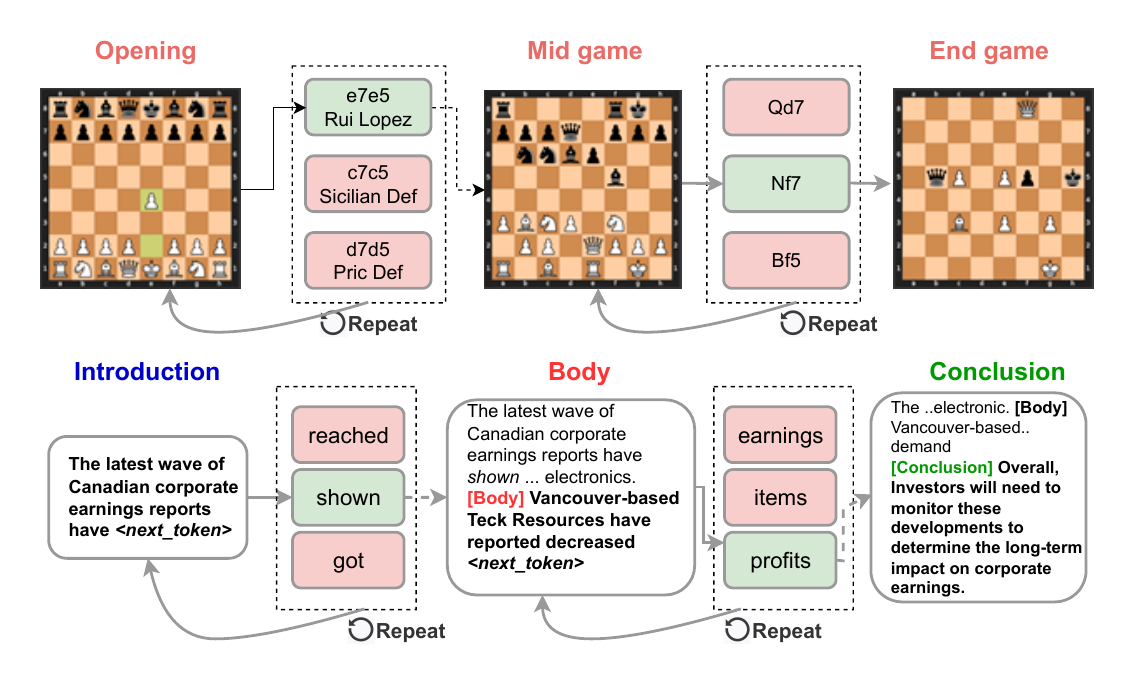}
    \caption{An illustration of the resemblance between chess and AI text generation. In chess, players select the optimal move from valid options given a board state; in text generation, LLMs similarly choose the next word/token from the vocabulary based on context. Both processes can be divided into three distinct segments, each serving a specific role in shaping the outcome of the game or the meaning of the text.}
    \label{fig_teaser_figure}
\end{figure}

\begin{figure*}[htbp]
    \centering
    \begin{subfigure}{0.3\textwidth}
        \centering
        \includegraphics[width=\linewidth]{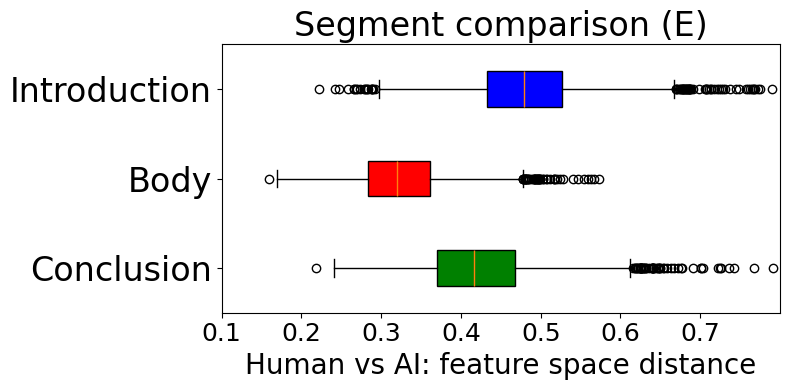}
        \caption{}
        \label{fig:segment_comparison_E}
    \end{subfigure}%
    \begin{subfigure}{0.3\textwidth}
        \centering
        \includegraphics[width=\linewidth]{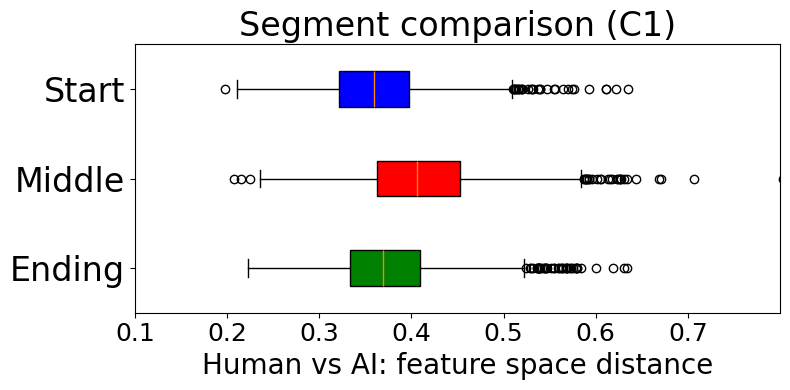}
        \caption{}
        \label{fig:segment_comparison_C1}
    \end{subfigure}%
    \begin{subfigure}{0.3\textwidth}
        \centering
        \includegraphics[width=\linewidth]{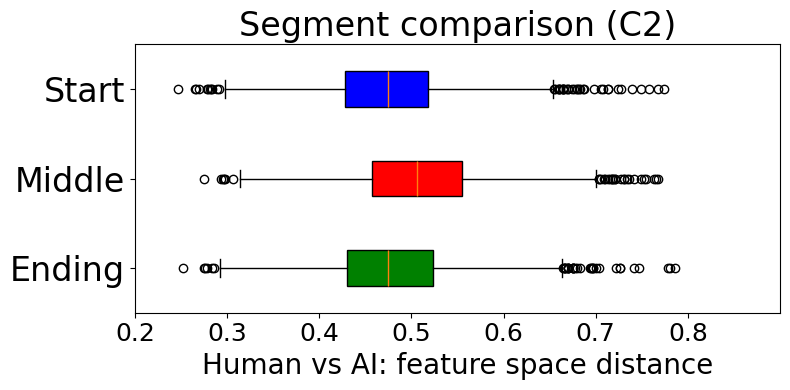}
        \caption{}
        \label{fig:segment_comparison_C2}
    \end{subfigure}
    \caption{Segment comparison results using LIWC \cite{boyd2022liwc} and WritePrint \cite{abbasi2008writeprints} features: \textbf{(a)} In the original setting \textit{(E)}, the body segment shows less difference between human and AI texts, likely due to its greater length.  Under length-controlled conditions: \textbf{(b)} \textit{C1} (equal segmentation) and \textbf{(c)} \textit{C2} (body matched to introduction/conclusion length), the body/middle segment exhibits the highest divergence.}
    \label{fig_segment_comparison}
\end{figure*}

Inspired by this transformation, we revisit the metaphor of chess to investigate a new frontier: understanding how human and AI-generated texts differ across \textit{segments}. In both chess and writing, structure matters. A chess match progresses through the opening, middlegame, and endgame, each demanding different levels of strategic reasoning. Likewise, written texts often follow a tripartite structure: an introduction to set the stage, a body to deliver core arguments, and a conclusion to synthesize insights. Chess opening and endgame moves are often heavily studied, analyzed, and codified into established theories for AI chess engines, like IBM DeepBlue \cite{campbell2002deepblue} or StockFish \cite{stockfish}. However, it is the dynamic middlegame where the true mastery of players is put to the test \cite{znosko1922middle}.
As Brian Christian \cite{christian2011most} explores in his book ``\emph{The Most Human Human}'', the middlegame represents the crucible where creativity, strategy, and adaptability separate humans from AI.

Just as in the middlegame of chess, one critical question arises: can LLMs move beyond following the typical opening and ending from their training data to navigate the fluid ``middlegame'' of text generation with the same linguistic ingenuity as humans? While recent studies have made substantial progress in distinguishing LLM-generated (AI text) from human-written text using stylometric features \cite{munoz2023contrasting, rosenfeld2024whose,guo2024benchmarking, reinhart2025llms}, thus \textit{checkmating} them, they often overlook the \textit{structural} context of the text. Do different text segments contribute differently to AI detection? And more importantly, do humans and LLMs exhibit similar patterns of stylistic variation across these segments? The answer has important implications, as limitations in this area could hinder their effectiveness in creative domains, while success would reinforce their role as versatile writing assistants.

Therefore, in this paper, we explore these questions through a comprehensive computational analysis of human and AI texts, focusing on three domains, news articles, essays, and emails, all of which naturally follow a structured format \cite{henry1997investigation,medvid2019essay,matruglio2020beyond}. Our dataset includes both human texts and generations from four prominent LLMs: ChatGPT (\textit{GPT-3.5}), PaLM (\textit{text-bison-001}), LLaMA2 (\textit{llama2-chat-7b}), and Mistral (\textit{mistral\_7b}). We introduce two core analyses:

\begin{enumerate}
    \item \textbf{Segment Comparison:} Do differences between human and AI texts vary across segments?
    \item \textbf{Source Comparison:} Do internal stylistic variation across segments differ between humans and AI texts ?
\end{enumerate}

Our findings are both surprising and insightful. While body segments initially appear more similar between human and AI texts (Figure \ref{fig_segment_comparison}), this is largely due to their greater length \cite{revesz2014laws}. In length-controlled settings, the body (or middle) consistently reveals the most significant differences. Moreover, it plays a dominant role in AI text detection. We also find that humans exhibit more variation across text segments than LLMs, reinforcing that LLMs tend to maintain a consistent stylistic fingerprint throughout. To further ground our analogy, we also analyze over 166K chess games to examine how human and AI players differ across game phases, showing that divergence peaks in the middlegame, the creative core of a match. Overall, our research sheds new light on the nuanced distinctions between human and AI text, offering a compelling step toward understanding the subtle yet defining elements that make human writing authentically human.

\begin{figure*}
    \centering
    \includegraphics[width=\linewidth]{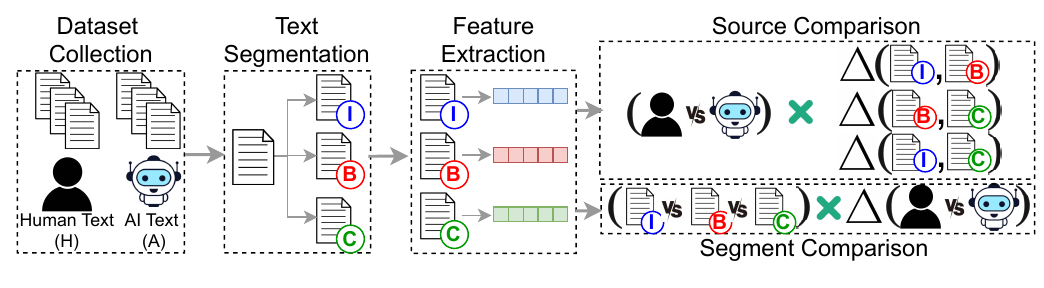}
    \caption{Overview of our methodology. Given a dataset of parallel human and AI texts, we divide each document into three segments and extract a comprehensive set of features from each segment. We perform statistical significance tests for \textbf{segment} and \textbf{source}  comparisons for each feature, considering all possible combinations.}
    \label{fig_methodology}
\end{figure*}

\section{Related Works}
\paragraph{Stylometry difference between human and AI text}
Stylometry features have long been effective in text classification and authorship analysis tasks, and can be proxies for creative \textit{chokepoints} in text  \cite{neal2017surveying}. With the growing availability of LLM-generated text datasets \cite{ dugan-etal-2024-raid, tripto2023hansen, verma2024ghostbuster}, recent research has applied these features to distinguish between human and AI text. For example, AI texts often differ from human writing in vocabulary diversity \cite{munoz2023contrasting}, distinctive word choices \cite{berriche2024unveiling}, formality \cite{al2024stylometric}, and rhetorical styles \cite{reinhart2025llms}. Therefore, several studies have leveraged linguistic features for AI text detection \cite{casal2023can, guo2024benchmarking, rosenfeld2024whose}, citing their explainability \cite{munoz2023contrasting} and strong statistical performance \cite{herbold2023large}. However, LLMs become increasingly adept at mimicking human writing styles, and their difference is narrowing \cite{toshevska2025llm}. 

\paragraph{AI text detection}
With the rapid advancement of LLMs, interest in detecting AI-generated text has surged across domains. Beyond stylometry-based methods, current detection approaches include fine-tuned models like the RoBERTa-based OpenAI Detector \cite{solaiman2019release}, GROVER \cite{zellers2019defending}, MAGE \cite{li-etal-2024-mage}, RADAR \cite{hu2023radar}, and LLM-DetectAIve \cite{abassy2024llm}, which use supervised learning on binary classification tasks (human vs. AI). In contrast, statistical and zero-shot detectors, such as DetectGPT \cite{mitchell2023detectgpt}, DetectLLM \cite{su2023detectllm}, GPT-who \cite{venkatraman2023gptwho}, and Binoculars \cite{hans2024binoculars} leverage distributional differences, often via perplexity, to offer more robust cross-domain performance. Commercial tools like GPTZero \cite{GPTzero}, Originality.ai \footnote{\url{https://originality.ai/ai-checker}}, and Turnitin’s AI detector \footnote{\url{https://www.turnitin.com/campaigns/clarity//}} also provide user-facing solutions. While many of these methods highlight important tokens for interpretability, they generally overlook which text segments contribute most to detection. By analyzing how different linguistic differences vary across text segments, our study offers a novel and necessary extension to the current literature, advancing the theoretical understanding and practical methodologies for AI text detection.

\section{Methodology}
Motivated by the chess middlegame analogy, we examine how human and AI texts differ across different segments. Figure \ref{fig_methodology} presents an overview of our methodological framework.

\subsection{Dataset creation}

We compile datasets from three domains (news articles, emails, and essays), each containing human-authored texts paired with corresponding LLM-generated versions. Our study includes four LLMs: ChatGPT (\textit{gpt-3.5-turbo}) from OpenAI, PaLM (\textit{text-bison-001}) from Google, LLaMA2 (\textit{llama2-chat-7b}) from Meta, and \textit{Mistral\_7b} from Mistral AI, representing both open-source and proprietary models. For the essay domain, we use the \textit{Persuade} corpus \cite{crossley2022persuade_corpus}, consisting of argumentative essays written by US students (grades 6-12) across different prompts. Our dataset includes approximately \textit{1700} human samples and corresponding LLM generations from the Kaggle competition \cite{king2023persuade_competition}. For news, we employ the Ghostbuster dataset \cite{verma2024ghostbuster}, which contains \textit{Reuters} articles and existing LLM generations (we generate missing samples using identical prompts). 

For the email domain, we draw on a curated subset of the Enron corpus \cite{klimt2004enron}, which still remains one of the most widely used and publicly available resources for authorship and stylistic analysis \citep{Tyo_Dhingra_Lipton_2022_AV, nini2024authorship}. Contemporary email datasets are scarce and often unsuitable, as many focus on spam detection, security leaks (e.g., Clinton emails \citep{shane2015hillary}), or bulk announcements (e.g., DBWorld \citep{filannino2011dbworld}), rather than personal correspondence. To ensure suitability, we perform extensive preprocessing: filtering for one-to-one internal emails to retain a personal tone, excluding automated, forwarded, or bulk messages, and removing emails with attachments. 
We also discard emails that are too short or excessively long, selecting only users with at least ten messages to ensure sufficient representation. This process yields a clean subset balancing realism with consistency. Using this data, we prompt LLMs to generate responses based on the original email’s header, sender/recipient information, and a concise content summary, with full prompt details provided in the Appendix \ref{appendix_prompts}. Table \ref{tab_dataset} summarizes key statistics across domains.

\begin{table}[h]
  \centering
  \resizebox{\columnwidth}{!}{
\begin{tabular}{c|c|cccc}
\hline
\textbf{\begin{tabular}[c]{@{}c@{}}Dataset\\ (Domain)\end{tabular}}                   & \textbf{Source} & \textbf{\# texts} & \textbf{\begin{tabular}[c]{@{}c@{}}Avg. \# \\ words\end{tabular}} & \textbf{\begin{tabular}[c]{@{}c@{}}Avg. \#\\ sentences\end{tabular}} & \textbf{\begin{tabular}[c]{@{}c@{}}I-B-C\\ ratio(\%)\end{tabular}} \\ \hline 
\multirow{2}{*}{\textbf{\begin{tabular}[c]{@{}c@{}}Reuter\\ (News)\end{tabular}}}     & Human           & 989               & 310.90                                                            & 10.98                                                                & 13-67-20                                                           \\ \cline{2-6} 
                                                                                      & AI              & 4741              & 288.05                                                            & 10.87                                                                & 15-57-28                                                           \\ \hline 
\multirow{2}{*}{\textbf{\begin{tabular}[c]{@{}c@{}}Enron\\ (Email)\end{tabular}}}     & Human           & 1632              & 173.34                                                            & 8.78                                                                 & 17-70-13                                                           \\ \cline{2-6} 
                                                                                      & AI              & 6289              & 144.61                                                            & 8.63                                                                 & 17-63-20                                                           \\ \hline 
\multirow{2}{*}{\textbf{\begin{tabular}[c]{@{}c@{}}Persuade\\ (Essays)\end{tabular}}} & Human           & 1717              & 269.93                                                            & 13.58                                                                & 18-60-22                                                           \\ \cline{2-6} 
                                                                                      & AI              & 3788              & 280.38                                                            & 13.71                                                                & 18-56-26                                                           \\ \hline
\end{tabular}
}
\caption{Key characteristics of human and AI texts across domains. Word and sentence counts per document are comparable between the two. I-B-C denotes the ratio of introduction (I), body (B), and conclusion (C) lengths in the original setting (Setting \textit{E}), with the body segment consistently longer than the others.}
  \label{tab_dataset}
\end{table}

\subsection{Text segmentation}
Segmenting text into introduction, body, and conclusion is inherently subjective \cite{hearst1994multi, aumiller2021structural}, as these sections often lack clear boundaries and vary significantly across writing styles, domains, and contexts. Manually annotating a large dataset would be prohibitively expensive and time-consuming. However, recent advances in LLMs have demonstrated strong performance in natural language understanding tasks, often achieving human-level performance \cite{thapa2023humans, michelmann2025large, sun2024head}. Therefore, we employ \textit{gemini-1.5-flash} (excluded from our authorship analysis to mitigate bias) to segment texts in our original setting (\textit{E}). Since body segments are typically longer \cite{henry1997investigation, raharjo2016generic}, we also explore length-controlled segmentation: in \textit{C1}, dividing texts into three equal parts, and in \textit{C2}, sampling a body portion matching the average length of the introduction and conclusion. In all settings, we ensure that the segments contain complete sentences to preserve semantic coherence and readability \cite{van1980interdisciplinary, graesser2003readers}.

Given the subjective nature of text segmentation, we show that our LLM-based approach is robust and well-aligned with alternative methods. We use the Segmentation Similarity Score \cite{fournier2012segmentation} (0 to 1, where 1 indicates identical segmentation) to evaluate text segmentation based on sentence counts. To validate our method, we segment a subset of 300 samples across all domains. Two human annotators (authors of this paper) provide manual segmentations to assess alignment with human perception, and we use \textit{GPT-4} to evaluate consistency between LLMs. Additionally, we fine-tune a BERT model on the human-segmented data to compare with standard computational techniques. As shown in Table \ref{tab_segmentation_result}, all comparisons yield segmentation similarity scores above 90\%, with no statistically significant differences ($\alpha$ = 0.05) among human-human, LLM-LLM, and LLM-human pairings. These results confirm that our LLM-based method, though not exact, reliably captures the structure of segmented text.

\begin{table}[]
\centering
\resizebox{\columnwidth}{!}{
\begin{tabular}{@{}ll|ll@{}}
\hline
\textbf{Dataset/Source} & \textbf{S} & \textbf{Judgement criteria} & \textbf{S}    \\ \hline
Persuade                & 0.96       & Gemini vs GPT4              & 0.93 \\
Enron                   & 0.90       & Gemini vs Human 1           & 0.91 \\
Reuter                  & 0.87       & Gemini vs Human 2           & 0.92 \\ \cline{1-2}
Human                   & 0.87       & GPT4 vs Human 1             & 0.92 \\ 
ChatGPT                 & 0.91       & GPT4 vs Human 2             & 0.91 \\
PaLM                    & 0.93       & Human 1 vs Human 2          & 0.94 \\
Llama-2                 & 0.93       & Gemini vs Finetuned BERT    & 0.92 \\
Mistral                 & 0.96       &                             &      \\ \hline
\end{tabular}
}
\caption{Segmentation Similarity Score (S) across datasets, LLMs, and criteria. Scores are higher for essays \& emails with clearer structure, but lower for news. AI texts are easier to segment than human texts. The similarity scores across humans, LLMs, human–LLM pairs, and LLM–computational methods are nearly identical, with no statistically significant differences.}
\label{tab_segmentation_result}
\end{table}


\subsection{Feature extractions}
We extract traditional stylometric feature sets such as LIWC (Linguistic Inquiry and Word Count), which provides psycholinguistic characteristics \cite{boyd2022liwc}, and Writeprint features, which capture an author's distinctive stylometric patterns \cite{abbasi2008writeprints}. Additionally, we examine how specific features vary across different segments and sources. Therefore, we include several individual lexical (vocabulary richness, readability), syntactic (part-of-speech tags, named entity tags,  stopwords distributions) opinion (formality, sentiment, subjectivity), contextual (text embedding), and text perplexity-related features, offering a comprehensive analysis of the text's stylistic and structural attributes (details in Appendix \ref{appendix_text_features}).


To use these features for \textbf{segment} and \textbf{source} comparison using statistical significance tests, we first define a difference measure, denoted as $\Delta$, between two feature values. Features are categorized as either scalar (e.g., vocabulary richness, readability, sentiment score) or distributional (e.g., POS-tag, stopwords, and LIWC distributions). For scalar features, we use absolute difference. For distributional features, we apply Jensen–Shannon Divergence (JSD) \cite{lin1991divergence}, a symmetric, bounded metric well-suited for comparing discrete probability distributions \cite{endres2003jsd}. For vector-based features not summing to one, such as perplexity scores and contextual embeddings, we use correlation distance and cosine distance, respectively. These capture relational and angular differences, making them appropriate for high-dimensional comparisons \cite{ruppert2004elements, huang2008similarity, turney2010frequency}.




\begin{table}[h]
\centering
  \resizebox{\columnwidth}{!}{
           \begin{tabular}{@{}l|l@{}}
\hline
\textbf{Criteria}                    & \textbf{Description}                         \\ \hline
\textbf{White vs Black}              & Human as white (53.08\%), AI as white (46.92\%)                     \\
\textbf{AI win \% as white}          & Win (71.36\%), Draw (5.29\%), Loss (23.35\%) \\
\textbf{AI win \% as black}          & Win (67.23\%), Draw (4.79\%), Loss (27.98\%) \\
\textbf{Elo ratings} & Human (1503-2433), AI (1557-2761)            \\ 
\textbf{Game category} & Blitz (29.71\%), Lighting (29.29\%), Standard (41\%) \\
\textbf{Move category} & Opening (28.31\%), Middle (29.23\%), End (42.46\%) \\
\textbf{Top 4 ECO codes} & A00(4.54\%), A45(4.09\%), D00(3.23\%), C50(2.37\%) \\
\hline
\end{tabular}
}
\caption{ Overview of the chess games analyzed in our study. The AI players generally have higher Elo ratings and win percentages compared to their human counterparts
in the dataset.}
\label{tab_chess_dataset}
\end{table}

\subsection{Statistical significance test}
\label{subsec_statistical_test}
 As we are interested in understanding how linguistic features differ between human  and AI  texts across textual segments (Introduction, Body, Conclusion), we design statistical significance tests with feature values extracted from each segment. Specifically, 
We conduct separate statistical tests for each linguistic feature.  Given two text sources \textbf{(Sources, \(H\): Human, \(A\): AI)} and three segments from each text \textbf{(Segments, \(I\): Introduction, \(B\): Body, \(C\): Conclusion)}, we define \(Z_x\) as an individual feature from segment \(x\) for source \(Z\).

For \textbf{source comparison} tests, we consider pairwise segments, \(x, y \in \{I, B, C\}\), compute their differences for human and AI texts, \(\Delta(H_x, H_y)\) and \(\Delta(A_x, A_y)\), respectively. We evaluate whether human cross-segment differences \(\Delta(H_x, H_y)\) are statistically greater than (\(>\)), less than (\(<\)), or comparable (\(\sim\)) to AI cross-segment differences \(\Delta(A_x, A_y)\), for specific pair of segments. Similarly, for  \textbf{segment comparison}, we compute the difference between human and AI texts for all three segments, \(\Delta(H_I, A_I)\), \(\Delta(H_B, A_B)\), and \(\Delta(H_C, A_C)\) to determine whether human-AI differences are statistically similar across segments. Details of the tests are mentioned in the Appendix \ref{appendix_statistical_test}.

\subsection{Chess dataset creation}
Since our study was motivated by the chess middlegame analogy, we conduct a concise yet systematic analysis of chess games to computationally explore whether these differences vary by phases. 
Using games from the Free Internet Chess Server (FICS) database\footnote{\url{https://www.ficsgames.org/download.html}}, we compile a dataset of ranked human vs AI games played between 2018 and 2020, selected due to the rise of AlphaZero \cite{silver2018alphazero} and the emergence of open-source AI chess bots \cite{mcilroy2020aligning}. We include only games between 30 and 100 moves, excluding short (due to early blunders or resignations) or excessively long games (repetitive moves). Table \ref{tab_chess_dataset} summarizes the final dataset of 166,738 games. We then segment each game into opening, middlegame, and endgame phases and extract features from chess moves in each segment (see Appendix \ref{appendix_chess_features} for details).

\begin{table*}[]
\centering

{\tiny
\begin{tabular}{ll|ccc|c}
\hline
 &  & \multicolumn{3}{c|}{\textbf{Source comparison}} &  \\ 
\multirow{-2}{*}{\textbf{Feature}} & \multirow{-2}{*}{\textbf{Dataset}} & \multicolumn{1}{l|}{\textbf{$\Delta$(I, B)}} & \multicolumn{1}{l|}{\textbf{$\Delta$(I, C)}} & \multicolumn{1}{l|}{\textbf{$\Delta$(B, C)}} & \multirow{-2}{*}{\textbf{\begin{tabular}[c]{@{}c@{}}Segment comparison\\ $\Delta$(H,A)\end{tabular}}} \\ \hline 

 & \textbf{Reuter} & \multicolumn{1}{c|}{\cellcolor[HTML]{CBCEFB}H\textgreater{}A} & \multicolumn{1}{c|}{$\sim$} & \cellcolor[HTML]{CBCEFB}H\textgreater{}A & \cellcolor[HTML]{CFFC99}B\textgreater{}C\textgreater{}I \\ 
 & \textbf{Enron} & \multicolumn{1}{c|}{$\sim$} & \multicolumn{2}{c|}{\cellcolor[HTML]{CBCEFB}H\textgreater{}A} & \multicolumn{1}{r}{$\sim$  \hspace{6mm} ($\dag$ $\ddag$)}   \\  
\multirow{-3}{*}{\textbf{\begin{tabular}[c]{@{}l@{}}Vocabulary\\ Richness\end{tabular}}} & \textbf{Persuade} & \multicolumn{1}{c|}{$\sim$} & \multicolumn{2}{c|}{\cellcolor[HTML]{CBCEFB}H\textgreater{}A} & \cellcolor[HTML]{CFFC99}B\textgreater{}C\textgreater{}I \\ \hline

 & \textbf{Reuter} & \multicolumn{3}{c|}{\cellcolor[HTML]{CBCEFB}H\textgreater{}A} & \multicolumn{1}{r}{\cellcolor[HTML]{FFCCC9}C\textgreater{}I\textgreater{}B \hspace{5.2mm} $\dag$ $\ddag$} \\ 
 & \textbf{Enron} & \multicolumn{1}{c|}{\cellcolor[HTML]{FFCE93}A\textgreater{}H} & \multicolumn{2}{c|}{\cellcolor[HTML]{CBCEFB}H\textgreater{}A} & \multicolumn{1}{r}{\cellcolor[HTML]{FFCCC9}I\textgreater{}C\textgreater{}B \hspace{7mm} $\dag$} \\ 
\multirow{-3}{*}{\textbf{\begin{tabular}[c]{@{}l@{}}Readability \\ Score\end{tabular}}} & \textbf{Persuade} & \multicolumn{3}{c|}{$\sim$} & \multicolumn{1}{r}{\cellcolor[HTML]{FFCCC9}C\textgreater{}I\textgreater{}B \hspace{5.2mm} $\dag$ $\ddag$}  \\ \hline

 & \textbf{Reuter} & \multicolumn{3}{c|}{$\sim$} & \multicolumn{1}{r}{\cellcolor[HTML]{FFCCC9}I$\sim$C\textgreater{}B \hspace{7.7mm}$\Diamond$}  \\  
 & \textbf{Enron} & \multicolumn{3}{c|}{\cellcolor[HTML]{FFCE93}A\textgreater{}H} & \multicolumn{1}{r}{C\textgreater{}B\textgreater{}I \hspace{8.5mm}$\ddag$ }\\ 
\multirow{-3}{*}{\textbf{\begin{tabular}[c]{@{}l@{}}Sentiment \\ Score\end{tabular}}} & \textbf{Persuade} & \multicolumn{3}{c|}{$\sim$} &  \multicolumn{1}{r}{\cellcolor[HTML]{FFCCC9}I\textgreater{}C\textgreater{}B \hspace{7.7mm}$\Diamond$}  \\ \hline

 & \textbf{Reuter} & \multicolumn{3}{c|}{\cellcolor[HTML]{CBCEFB}H\textgreater{}A} &  
 \multicolumn{1}{r}{\cellcolor[HTML]{FFCCC9}I$\sim$C\textgreater{}B  \hspace{6mm}$\dag$ $\ddag$}  \\ 
 & \textbf{Enron} & \multicolumn{3}{c|}{\cellcolor[HTML]{CBCEFB}H\textgreater{}A} & \multicolumn{1}{r}{\cellcolor[HTML]{FFCCC9}I$\sim$C\textgreater{}B  \hspace{8.2mm}$\dag$} \\  
\multirow{-3}{*}{\textbf{\begin{tabular}[c]{@{}l@{}}Formality Score \&\\  Content Similarity\\ (same results)\end{tabular}}} & \textbf{Persuade} & \multicolumn{3}{c|}{\cellcolor[HTML]{CBCEFB}H\textgreater{}A} &  \multicolumn{1}{r}{\cellcolor[HTML]{FFCCC9}I$\sim$C\textgreater{}B  \hspace{6mm}$\dag$ $\ddag$}  \\ \hline

 & \textbf{Reuter} & \multicolumn{3}{c|}{$\sim$} & \cellcolor[HTML]{CFFC99}B\textgreater{}I\textgreater{}C \\ 
 & \textbf{Enron} & \multicolumn{1}{c|}{\cellcolor[HTML]{CBCEFB}H\textgreater{}A} & \multicolumn{1}{c|}{\cellcolor[HTML]{FFCE93}A\textgreater{}H} & \cellcolor[HTML]{CBCEFB}H\textgreater{}A & \multicolumn{1}{r}{\cellcolor[HTML]{FFCCC9}C\textgreater{}I\textgreater{}B \hspace{7.9mm}$\Diamond$ }\\ 
\multirow{-3}{*}{\textbf{\begin{tabular}[c]{@{}l@{}}Perplexity \\ Scores\end{tabular}}} & \textbf{Persuade} & \multicolumn{3}{c|}{$\sim$} & \cellcolor[HTML]{CFFC99}B\textgreater{}I$\sim$C \\ \hline

 & \textbf{Reuter} & \multicolumn{3}{c|}{\cellcolor[HTML]{CBCEFB}H\textgreater{}A} & \multicolumn{1}{r}{\cellcolor[HTML]{FFCCC9}I\textgreater{}C\textgreater{}B \hspace{7mm} $\Diamond$} \\ 
 & \textbf{Enron} & \multicolumn{3}{c|}{\cellcolor[HTML]{CBCEFB}H\textgreater{}A} & \multicolumn{1}{r}{\cellcolor[HTML]{FFCCC9}I$\sim$C\textgreater{}B \hspace{7mm} $\Diamond$} \\ 
\multirow{-3}{*}{\textbf{\begin{tabular}[c]{@{}l@{}}Parts of Speech\\ Tags Distribution\end{tabular}}} & \textbf{Persuade} & \multicolumn{3}{c|}{\cellcolor[HTML]{CBCEFB}H\textgreater{}A} & \multicolumn{1}{r}{\cellcolor[HTML]{FFCCC9}I\textgreater{}C\textgreater{}B \hspace{7mm} $\Diamond$}\\ \hline 

 & \textbf{Reuter} & \multicolumn{1}{c|}{\textbf{$\sim$}} & \multicolumn{2}{c|}{\cellcolor[HTML]{CBCEFB}H\textgreater{}A} & \multicolumn{1}{r}{\cellcolor[HTML]{FFCCC9}C\textgreater{}I\textgreater{}B \hspace{7.2mm} $\Diamond$ }\\ 
 & \textbf{Enron} & \multicolumn{1}{c|}{\textbf{$\sim$}} & \multicolumn{1}{c|}{\cellcolor[HTML]{CBCEFB}H\textgreater{}A} & $\sim$ & $\sim$ \\ 
\multirow{-3}{*}{\textbf{\begin{tabular}[c]{@{}l@{}}Named Entity\\ Tags Distribution\end{tabular}}} & \textbf{Persuade} & \multicolumn{1}{c|}{\textbf{$\sim$}} & \multicolumn{2}{c|}{\cellcolor[HTML]{CBCEFB}H\textgreater{}A} & $\sim$ \\ \hline
\end{tabular}
}
\caption{Statistical significance test results in the original experimental setting \textit{(E)}.
\textbf{Source Comparison:} $\Delta$(I, B) represents the difference in a given feature between the Introduction (I) and Body (B) for both human and AI texts. \hlc[CBCEFB]{Violet (H \textgreater{} A)} indicates that this difference is significantly greater in human texts, while \hlc[FFCE93]{orange (A \textgreater{} H)} denotes the opposite and ($\sim$) indicates no statistically significant difference.
\textbf{Segment Comparison:} $\Delta$(H, A) captures the feature difference between human and AI texts within a specific segment (I, B, or C). \hlc[CFFC99]{Green} highlights cases where the body segment shows a significantly greater difference than the introduction or conclusion, while \hlc[FFCCC9]{red} marks the opposite. ($\sim$) indicates no significant difference across segments.
The symbols ($\dag$) and ($\ddag$) denote cases where the body segment shows higher differences in the length-controlled settings \textit{C1} and \textit{C2}, respectively. The $\Diamond$ symbol indicates no significant segmental difference in both \textit{C1} and \textit{C2}. Cells without symbols represent cases where the original setting (E) aligns with both length-controlled settings.}
\label{tab_statistical_result}
\end{table*}
\section{Results}
We present our findings on \textbf{segment} and \textbf{source} comparisons across different experimental settings, identify which text segment contributes most to AI text detection, and explore whether similar segmental differences exist between human and AI chess players.
\subsection{Segment and source comparison results}
We conduct a comprehensive analysis of individual features across all possible combinations to evaluate both \textbf{segment} and \textbf{source} comparisons, with key findings summarized in Table \ref{tab_statistical_result}. In the original experimental setting (\textit{E}), \textbf{segment comparison} reveals that the body segment exhibits less distinction between human and AI texts compared to the introduction and conclusion. However, this lower contrast is due to the body’s greater length, which can dilute syntactic features like POS-tag or named entity distributions and flatten opinion-based features such as sentiment or formality through averaging. The extended length also allows AI text to align more easily with human content in the body segment. Nevertheless, length-independent features like vocabulary richness and perplexity indicated higher differences in the body.

In the length-controlled experiments (\textit{C1} and \textit{C2}) settings, stylometric (e.g., LIWC, Writeprints) and linguistic features (e.g., vocabulary richness, readability, sentiment) show higher differences in the body/middle segment. When segment lengths are normalized, several features show no statistically significant differences across segments. Given that the body segment typically hosts the core arguments, elaboration, and creativity in writing \cite{medvid2019essay}, our findings suggest that while LLMs may mimic surface-level structure, they struggle to replicate the nuanced, adaptive strategies humans employ in this more demanding segment, as validated through human vs. AI text detection in the following subsection.

In the \textbf{source comparison}, our findings show human texts exhibit higher cross-segment variation than AI text, offering an innovative lens to differentiate between the two. While prior studies \cite{guo2023HC3,munoz2023contrasting} have shown that AI texts tend to be more structurally consistent and formal, our analysis uncovers how this consistency manifests across segments. LLMs inherently prefer structured text generation, often incorporating a distinct introduction, body, and conclusion boundary, leading to smoother transitions and uniform distribution of content, named entities, and POS tags across segments. In contrast, human writers tend to modulate their linguistic fingerprints between segments, a trait not yet replicated by AI. Additional analysis on individual LLM behavior can be found in the Appendix \ref{appendix_individual_llm_result}.

\begin{table*}[]
\centering
\resizebox{2\columnwidth}{!}{
\begin{tabular}{l|l|lllll|l}
\hline
\textbf{Dataset}    
& \textbf{Criteria}         & \textbf{GPT Zero} & \textbf{MAGE}   & \textbf{RADAR}  & \textbf{Binoculars} & \textbf{GPT-Who} & \textbf{Finetuned Bert} \\ \hline
\multirow{4}{*}{\begin{tabular}[c]{@{}l@{}}Reuter\\ (News)\end{tabular}}    & \textbf{Total text}       & 0.84              & 0.75            & 0.77            & 0.91                & 0.82             & 0.96                    \\
                                                                            & \textbf{Voting}           & 0.83 (↓1.19\%)    & 0.51 (↓45.74\%) & 0.72 (↓6.49\%)  & 0.85 (↓6.59\%)      & 0.82 (↓1.2\%)    & 0.97 (↑1.04\%)          \\
                                                                            & \textbf{Body only}        & 0.85 (↑1.19\%)    & 0.76 (↑1.33\%)  & 0.81 (↑5.19\%)  & 0.84 (↓7.69\%)      & 0.84 (↑2.44\%)   & 0.94 (↓2.08\%)          \\
                                                                            & \textbf{Intro+conclusion} & 0.76 (↓9.52\%)    & 0.62 (↓17.33\%) & 0.77 (↓0\%)     & 0.77 (↓15.38\%)     & 0.79 (↓3.66\%)   & 0.93 (↓3.12\%)          \\ \hline
\multirow{4}{*}{\begin{tabular}[c]{@{}l@{}}Enron\\ (Emails)\end{tabular}}   & \textbf{Total text}       & 0.62              & 0.78            & 0.82            & 0.73                & 0.77             & 0.98                    \\
                                                                            & \textbf{Voting}           & 0.61 (↓1.61\%)    & 0.78 (↑0.0\%)   & 0.73 (↓10.98\%) & 0.71 (↓2.74\%)      & 0.85 (↑10.39\%)  & 0.96 (↓2.04\%)          \\
                                                                            & \textbf{Body only}        & 0.71 (↑14.52\%)   & 0.72 (↓7.69\%)  & 0.79 (↓3.36\%)  & 0.74 (↑1.37\%)      & 0.78 (↑1.3\%)    & 0.93 (↓5.1\%)           \\
                                                                            & \textbf{Intro+conclusion} & 0.55 (↓11.29\%)   & 0.7 (↓10.26\%)  & 0.74 (↓9.76\%)  & 0.68 (↓6.85\%)      & 0.75 (↓2.6\%)    & 0.96 (↓2.04\%)          \\ \hline
\multirow{4}{*}{\begin{tabular}[c]{@{}l@{}}Persuade\\ (Essay)\end{tabular}} & \textbf{Total text}       & 0.94              & 0.94            & 0.79            & 0.82                & 0.83             & 0.99                    \\
                                                                            & \textbf{Voting}           & 0.9 (↓4.26\%)     & 0.75 (↓20.21\%) & 0.64 (↓18.99\%) & 0.86 (↑4.88\%)      & 0.8 (↓3.61\%)    & 0.97 (↓2.02\%)          \\
                                                                            & \textbf{Body only}        & 0.88 (↓6.38\%)    & 0.82 (↓12.77\%) & 0.67 (↓15.19\%) & 0.84 (↑2.44\%)      & 0.82 (↓1.2\%)    & 0.96 (↓3.03\%)          \\
                                                                            & \textbf{Intro+conclusion} & 0.89 (↓5.32\%)    & 0.73 (↓22.34\%) & 0.61 (↓22.78\%) & 0.78 (↓4.88\%)      & 0.75 (↓9.64\%)   & 0.96 (↓3.03\%)         \\ \hline
\end{tabular}
}
\caption{AI text detection results (original setting \textit{E}). Each cell value represents the F1 score of various detection methods, with higher scores indicating better performance. The results are presented across multiple datasets and evaluated using different criteria to assess how different segments can contribute to AI text detection.}
\label{tab_ai_detection}
\end{table*}

\subsection{Checkmating AI text: which segment reveals its origins?}
To explore how different text segments contribute to AI text detection, we evaluate a suite of prominent detectors: GPT-Zero \cite{GPTzero}, MAGE \cite{li-etal-2024-mage}, Radar \cite{hu2023radar}, Binocular \cite{hans2024spotting}, GPT-Who \cite{venkatraman2024gpt}, and a fine-tuned BERT classifier (detailed in the Appendix \ref{appendix_ai_text_detect}). Our primary goal is to understand the relative importance of the introduction, body, and conclusion in distinguishing human and AI text. Accordingly, we apply each detector to the total text, individual segments, and a combined introduction \& conclusion segment. We also test a simple voting mechanism across the three segments. Results are summarized in Table \ref{tab_ai_detection}.

\begin{figure}[htbp]
    \centering
    \begin{subfigure}{0.25\textwidth}
        \centering
        \includegraphics[width=\linewidth]{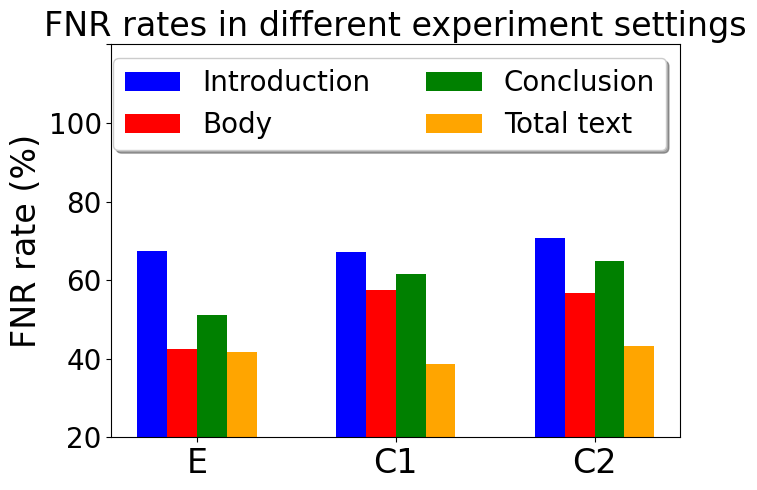}
    \end{subfigure}%
    \begin{subfigure}{0.25\textwidth}
        \centering
        \includegraphics[width=\linewidth]{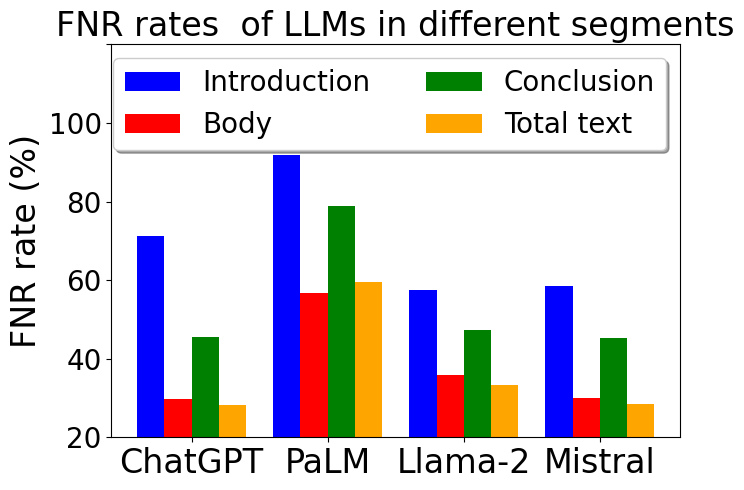}
    \end{subfigure}
    \caption{Comparison of False Negative Rates (FNR) in different experimental  settings \& datasets. Lower value indicates this segment contributes more in detection. }
    \label{fig:fnr_rate_exp_dataset}
\end{figure}

Overall, using the entire text yields the highest detection performance across most domains, except for the email. It aligns with the nature of email writing: introductions and conclusions often include formulaic greetings or closing remarks, while the body contains the most meaningful content. Across all domains, the body consistently plays a dominant role in AI text detection, outperforming both the introduction and conclusion, even when combined. Interestingly, the voting mechanism across segments fails to improve performance, likely due to redundancy or the overwhelming influence of the body segment. Notably, fine-tuned classifiers consistently benefit from analyzing the complete text, as they leverage more data during training. Appendix \ref{appendix_ai_text_detect} provides the AI text detection results for other experimental settings.

\begin{figure}[htbp]
    \centering
    \begin{subfigure}{0.24\textwidth}
        \centering
        \includegraphics[width=\linewidth]{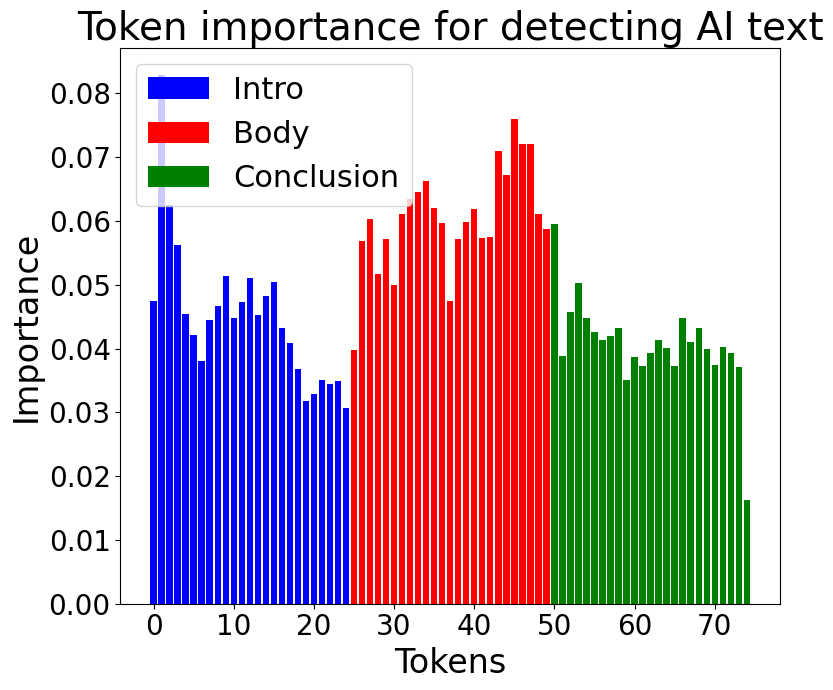}
        \caption{}
    \end{subfigure}%
    \begin{subfigure}{0.26\textwidth}
        \centering
        \includegraphics[width=\linewidth]{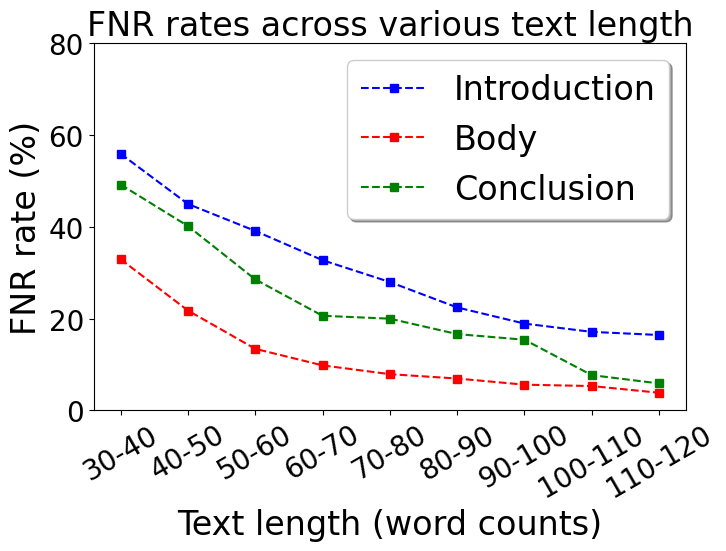}
        \caption{}
    \end{subfigure}
    \caption{\textbf{(a)} Average importance from each token for identifying AI-generated text, showing higher contributions from tokens in the body segment. \textbf{(b)} False Negative Rate (FNR) decreases as the word count of a given segment increases. Even when the introduction, body, and conclusion are around the same length, the
body segment has a lower FNR, making it stand out from the introduction and conclusion.}
    \label{fig_importance_line_chart}
\end{figure}

To account for the body segment’s longer length in the original setting \textit{(E)}, we assess detection performance using \textit{False Negative Rate} (FNR), the proportion of AI text misclassified as human, across all settings \& datasets (Figure \ref{fig:fnr_rate_exp_dataset}). A lower FNR indicates better detector performance, as the text is more easily identified as LLM-generated, making it more distinguishable from human text. Conversely, a higher FNR suggests that the text closely resembles human writing, causing the detector to struggle to label it as AI text. Consistently, the body segment yields the lowest FNR, suggesting that it is more distinguishable from human text than the introduction or conclusion. Prior work \cite{huang2024authorship,wu2024detectrl} shows that longer texts generally improve detection, a trend we confirm in (Figure \ref{fig_importance_line_chart}), where FNR declines as text length increases. Yet, within comparable length ranges, the body segment still exhibits the lowest FNR. To quantify which parts of the text contribute most to being flagged as AI text, we use Integrated Gradients \citep{sundararajan2017axiomatic} to estimate token-level importance in our fine-tuned BERT classifier. For each correctly predicted sample, we compute the gradient of the model's output with respect to each token and normalize the resulting attribution scores to obtain a list of token importances. We then divide each sample into three equal-length segments: start, middle, and end (mirroring our $C1$ segmentation strategy to minimize length confounds), and average normalized importance scores within each segment. These scores are then aggregated across all samples to produce a final token importance profile for each segment. As shown in Figure 5 (left), we find that the middle segment consistently receives higher attribution, suggesting that it plays a more decisive role in distinguishing AI-generated text from human-written text.


\begin{table}[h]
\centering
\resizebox{\columnwidth}{!}{
\begin{tabular}{@{}l|ll|ll|ll@{}}
\hline
\textbf{Dataset}  & \textbf{MAGE} & \textbf{MAGE+}               & \textbf{RADAR} & \textbf{RADAR+}              & \textbf{Binocular} & \textbf{Binocular+}          \\ \hline
\textbf{Reuter}   & 0.85          & \cellcolor[HTML]{CFFC99}0.87 & 0.69           & \cellcolor[HTML]{CFFC99}0.87 & 0.68               & \cellcolor[HTML]{CFFC99}0.91 \\
\textbf{Persuade} & 0.86          & \cellcolor[HTML]{CFFC99}0.88 & 0.84           & \cellcolor[HTML]{CFFC99}0.85 & 0.89               & \cellcolor[HTML]{CFFC99}0.90 \\
\textbf{Enron}    & 0.88          & \cellcolor[HTML]{FFCCC9}0.81 & 0.82           & \cellcolor[HTML]{FFCCC9}0.7  & 0.57               & \cellcolor[HTML]{CFFC99}0.65 \\ \hline
\end{tabular}
}
\caption{Cross-segment feature differences enhance the performance of base detectors in identifying AI text from human-AI text pairs. \hlc[CFFC99]{Green} cells indicate improved performance when using cross-segment variation instead of detector confidence scores, while \hlc[FFCCC9]{Red} cells indicate decreased performance.}
\label{tab_cross_segment_in_ai_text}
\end{table}

Finally,  cross-segment variation between human and AI texts (\textbf{source comparison} results)  prompts us to explore its utility in AI text detection. We frame the task as identifying the AI text from a given (human, AI) pair. When existing detectors assign the same label to both texts, rather than relying solely on their confidence scores (denoted as \textit{detector\_name}), we use the cross-segment variation (based on the C1 setting, which splits text into three equal parts and is more practical for real-world use) as the deciding factor (\textit{detector\_name+}). This simple yet effective strategy improves detection accuracy across most detectors and datasets (Table \ref{tab_cross_segment_in_ai_text}), demonstrating that cross-segment variation offers a promising new lens for AI text detection.

\subsection{Human and AI chess moves comparison}
\begin{figure}[htbp]
    \centering
    \begin{subfigure}{0.25\textwidth}
        \centering
        \includegraphics[width=\linewidth]{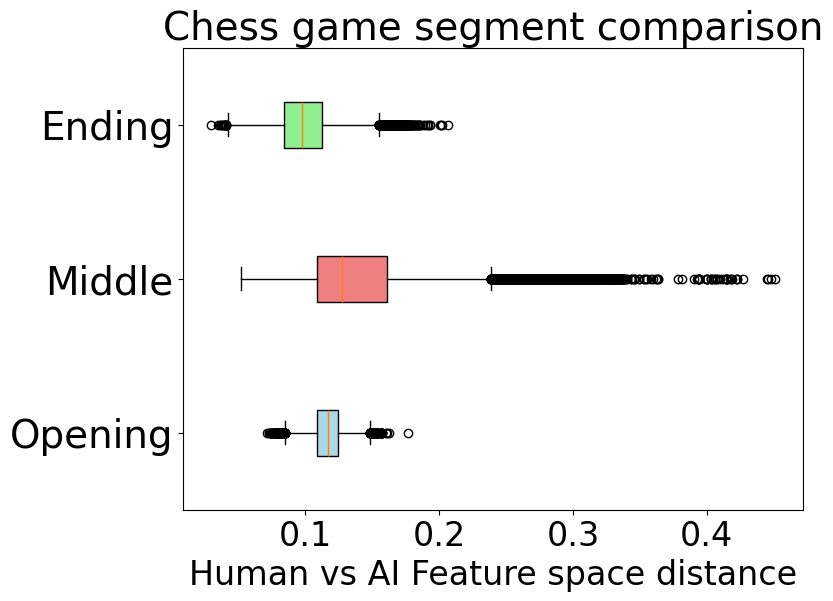}
        \caption{}
    \end{subfigure}%
    \begin{subfigure}{0.25\textwidth}
        \centering
        \includegraphics[width=\linewidth]{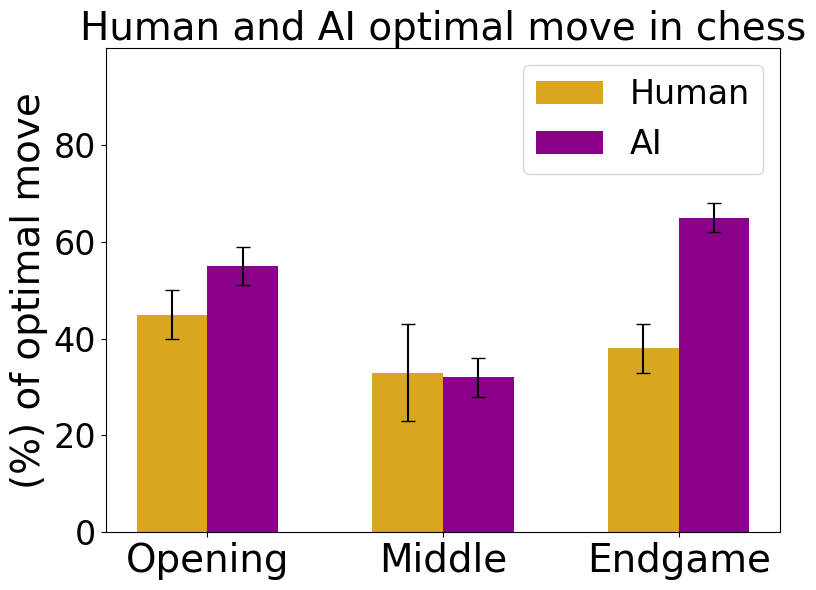}
        \caption{}
    \end{subfigure}
    \caption{\textbf{(a)} The middlegame exhibits the most significant divergence between human and AI players. \textbf{(b)} AI players outperform humans in optimal move percentage during the opening and endgame, but the difference is not statistically significant in the middlegame.}
    \label{fig_chess_results}
\end{figure}

As our study was inspired by the chess middlegame analogy,  We also investigate whether the  differences between human and AI players emerge most noticeably in the middlegame. To quantify these differences, we calculate the JSD distance between the feature sets of human and AI moves across the opening, middlegame, and endgame phases. As shown in Figure \ref{fig_chess_results}, the middlegame exhibits a statistically significant ($\alpha$ = 0.05) increase in JSD, indicating higher divergence during this phase. Moreover, the middlegame shows a broader spread of JSD values, reflecting higher variability in how humans and AI play diverges. We further compute Jaccard similarity over unique move patterns, represented by distinct Standard Algebraic Notation (SAN) moves exhibited by a player and observe lower overlap in the middlegame compared to the opening and endgame, reinforcing that this phase carries the most distinction. These findings echo our \textbf{text segment comparison} results, where the body or ``middlegame'' segment also reveals the highest differences between humans and AI. 

\begin{figure}[htbp]
    \centering
    \begin{subfigure}{0.25\textwidth}
        \centering
        \includegraphics[width=\linewidth]{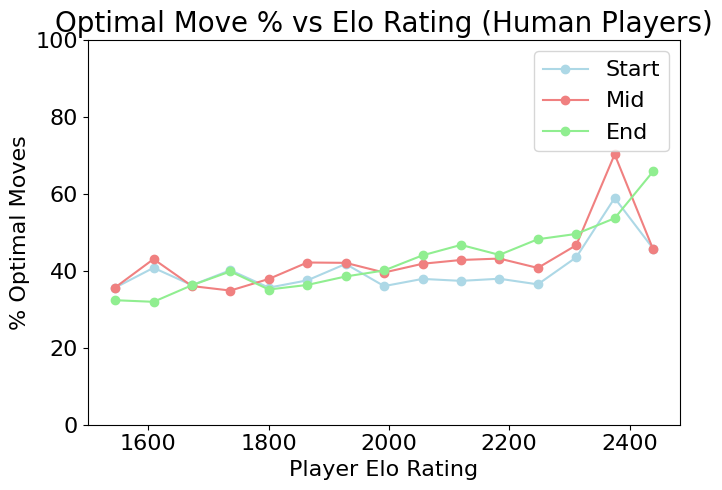}
    \end{subfigure}%
    \begin{subfigure}{0.24\textwidth}
        \centering
        \includegraphics[width=\linewidth]{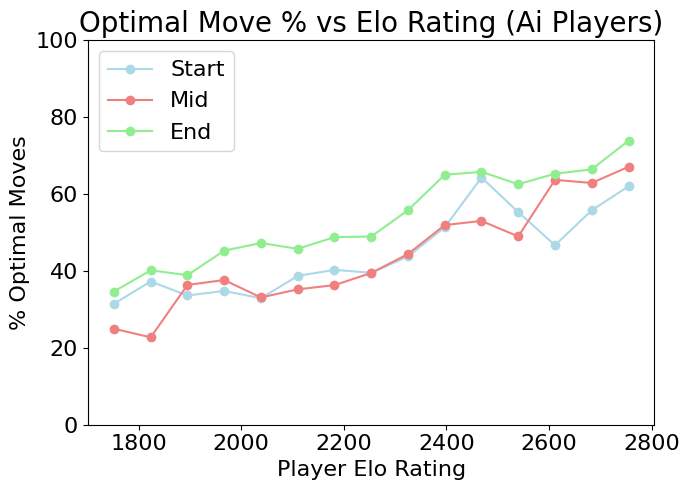}
    \end{subfigure}
    \caption{Percentage of optimal moves across game positions for human and AI players, plotted against player Elo rating. Each point represents the mean percentage of optimal moves within the corresponding Elo bin.}
    \label{fig_chess_results_elo}
\end{figure}

Finally, we analyze the percentage of optimal moves and win probability using the Stockfish game engine \cite{stockfish} for each move. As expected, AI players achieve higher optimal move rates and win probabilities, particularly in the endgame phase. AI engines often perform exceptionally well in endgames due to their access to precomputed endgame scenarios, which provide exact move sequences for optimal play to ensure victory. These tablebases \citep{thompson1986tablebases} are derived from exhaustive analysis rather than historical data and offer perfect information, giving AI engines a decisive advantage in such positions, an advantage human players, regardless of skill level, typically do not possess.
Additionally, we observe that the percentage of optimal moves generally increases with Elo rating, but at a steeper rate for AI players than for humans (Figure \ref{fig_chess_results_elo}). Within the same Elo range, AI players also make more optimal moves in the endgame compared to the middlegame, whereas for humans, performance remains relatively stable across these phases. While differences in play style between humans and AI across game segments \citep{christian2011most, mcilroy2021detecting} motivated part of our analysis, this was not the primary focus of the study, and we therefore refrained from a deeper investigation. Nonetheless, these findings align with and further validate our broader observations on segment-level distinctions, and we believe this direction merits dedicated exploration in future work.

\section{Discussion}
In this section, we highlight key findings that reinforce our central claim, offer valuable insights into human creativity, and demonstrate the broader applicability of our results.

\paragraph{Text length matters}
We find that LLMs' ability to replicate human stylometry and linguistic features is influenced by text length. Initially, the body segment appears more similar to human text due to its greater length. Longer texts also yield higher AI text detection accuracy, aligning with prior studies \cite{liu2020influence, liu2024study_text_length, baillargeon2024assessing, jeon2021countering}, which show improved classification and higher similarity scores in lengthier samples \cite{klaussner2015finding, papcke2023stylometric}. Therefore,  LLMs can better approximate human writing when given the chance to generate more tokens, as they have more room to establish consistent stylistic patterns, an insight critical to understanding and detecting AI text.

\paragraph{Distribution vs. textual divergence}
Our study offers a comprehensive view of how well LLMs replicate different linguistic features. 
LLMs consistently excelled at replicating the features that do not rely on word orders in sentences but instead depend on overall word choices, such as pos-tags, stopword distributions, or readability scores, showing no observable statistical differences with humans across experiments. In contrast, for features that capture the continuous flow of text, such as token-level perplexity or content change through that text, human and AI texts exhibited significant differences across experimental conditions. These insights can assist platforms like Turnitin, Grammarly, or Originality to integrate flow-based stylometric checks for AI text detection.


\paragraph{Body segment: more interesting for Human-AI text distinction}
While a longer body segment makes human and AI texts appear more stylistically similar for that segment, body/middle consistently shows higher divergence in length-controlled settings. Additionally, AI-generated introductions and conclusions yield higher false negative rates, suggesting detectors perceive them as more human-like. Token importance further confirms the body segment’s superior discriminatory power. Thus, when distinguishing between human and AI texts, the body segment offers the most revealing starting point.

\paragraph{Cross-segment variation as a signal for AI text detection}
Our \textbf{source comparison} shows that cross-segment linguistic and contextual differences are consistently more pronounced in human texts than in AI-generated ones. It suggests that LLMs maintain a uniform writing style across segments, while humans naturally vary their linguistic patterns throughout a text. Importantly, we find that leveraging these cross-segment stylometric differences as a secondary signal can enhance the performance of existing AI text detectors, highlighting a promising new direction for detection strategies.

\section{Conclusion}
Our paper offers a novel perspective by identifying subtle differences between human and AI texts across specific text segments, an area that has remained largely overlooked. Drawing parallels from chess game phases, we conduct a thorough evaluation of linguistic features, analogous to chess ``chokepoints''  and explore how they vary in each segment between AI and human text. Our experimental design and detailed segment-wise analysis offer robust insights into LLMs' strengths and limitations in mimicking human text. Overall, our findings highlight the pivotal role of the body segment in distinguishing AI from human text and propose that cross-segment feature differences may serve as a novel and valuable characteristic for AI text detection. In future, we aim to extend our findings to other domains and contribute to responsible LLM usage to ensure accurate outputs across all text segments.

\section*{Limitations}
While this study presents new findings in differentiating between human and AI text, inspired by chess game dynamics,  there are some limitations to acknowledge. First, the scope of our analysis is restricted to three domains and texts from four LLMs. Additionally, the AI texts are collected from existing datasets that used generic prompts, which may affect the generalization of our findings to other domains, models, or prompting techniques. Secondly, dividing a text into introduction, body, and conclusion is inherently subjective, and while we show that an LLM can perform this segmentation, demonstrating alignment with human judgment, alternative approaches may yield different results. Moreover, not all domains, such as creative writing or social media posts, naturally follow a tripartite structure. Thus, applying our framework to such cases will require special attention. Despite these constraints, our study makes a substantial contribution by exploring human-AI text distinctions from a novel angle and can inform ongoing AI text detection research.

\section*{Ethical Considerations}
Our study raises important ethical considerations regarding the responsible development, evaluation, and deployment of Large Language Models (LLMs). By analyzing segment-level distinctions between human and AI-generated texts, our goal is not to stigmatize AI use in writing but to promote transparency and accountability in its application. The insights from this research intend to strengthen detection mechanisms that help prevent misuse, such as academic dishonesty, misinformation, or deceptive authorship, while also informing the development of more interpretable and aligned LLMs. All AI-generated texts used in this study were created under controlled, non-deceptive conditions or collected from existing public datasets, and no personal, sensitive, or private human data was used. As detection technologies advance, it remains crucial to balance innovation with privacy, avoid over-surveillance, and ensure that such tools are not misused to unjustly penalize legitimate human writing.

\section*{Acknowledgments}
This work was in part supported by U.S. National Science Foundation (NSF) awards \#1934782 and \#2114824. Some of the research results were obtained using computational resources provided by National Artificial Intelligence Research Resource (NAIRR) award \#240336 and complimentary API support by the GPTZero.
D. Lee has a financial interest in the GPTZero, which is being appropriately managed by Penn State University.

\bibliography{main}

\appendix

\clearpage

\section{Prompt engineering}
\label{appendix_prompts}
While we primarily use human and AI text in various domains from existing datasets, we also employ LLMs for missing data generation and text segmentation. As mentioned, we select \textit{GPT-3.5} (OpenAI), PaLM \textit{text-bison-001} (Google), \textit{LLaMA 2-Chat-7B} (Meta), and \textit{Mistral-7B} (Mistral AI) as our LLMs. Several data were missing in the original datasets collected from  \cite{verma2024ghostbuster} or \cite{king2023persuade_competition}. 
For example, Reuters news articles from any Google model were unavailable in the original \textit{Ghostbuster} dataset \cite{verma2024ghostbuster}. So, we generated them using \textit{text-bison-001} using identical prompts from the original paper \cite{verma2024ghostbuster}. Similarly, for the email dataset, we generate AI text from all four LLMs, as only human-written emails are available in the Enron corpus \cite{klimt2004enron}.
For segmentation, we use \textit{Gemini-1.5-Flash} (Google) and \textit{GPT-4} (OpenAI), which are distinct from the models used for text generation in our study. Proprietary models from Google and OpenAI are accessed via their official APIs, while open-source models from Meta and Mistral are sourced from their stable weights on Hugging Face. Across all settings, we use \textbf{top\_p} = 0.95 and \textbf{temperature} = 0.9 to maintain consistency. However, it is important to note that even with identical prompts and hyperparameters, LLM outputs are not entirely deterministic.

\paragraph{Prompt for news data}
\begin{center}
\small
    \fbox{%
    \begin{minipage}{\linewidth}
        \texttt{Suppose You are <reporter\_name>, a news reporter in Reuter. Write a news article in <original\_word\_count> words with the following headline (output news text only, do not include headline):\\ \textbf{<original\_headline>}}
        
    \end{minipage}
}
\end{center}

\paragraph{Prompt for email data}
\begin{center}
\small
    \fbox{%
    \begin{minipage}{\linewidth}
        \texttt{Create an email (only the email body) as an Enron employee <sender\_name> to <receiver\_name> around <original\_word\_count> words based on the subject: <original\_email\_header>.
The summary of the original email is as follows.\\
\textbf{<original\_email\_summary>}}
    \end{minipage}
}
\end{center}

\paragraph{Prompt for text segmentation}
\begin{center}
\small
    \fbox{%
    \begin{minipage}{\linewidth}
        \texttt{You are advanced in essay understanding and writing. Given the following text you need to divide it into three parts: introduction, main body and conclusion. For each part, only copy relevant portion from the original text. Do not use any other formatting. \\
\{Introduction\}:the intro goes here \\
 \{Body\}:the main body goes here \\
 \{Conclusion\}:the conclusion goes here \\
The text is as follows: \\ \textbf{<original\_text>}}

    \end{minipage}
}
\end{center}

\begin{table*}[h]
\centering
  \resizebox{2\columnwidth}{!}{
 \begin{tabular}{@{}l|l|l|l|l@{}}
\toprule
\textbf{Opening conditions}                                                         & \textbf{reasonings}                                                                                                   & \textbf{Mid game}                                                                                                                            & \textbf{End game conditions}                                                                                                                       & \textbf{reasonings}                                                                                                        \\ \midrule
\begin{tabular}[c]{@{}l@{}}\# of moves \textless{}= 16\\ \\ \textbf{OR}\end{tabular}         & \begin{tabular}[c]{@{}l@{}}All classic chess openings are done in mostly\\  16 moves \cite{horowitz1986win}\end{tabular}            & \multirow{3}{*}{\begin{tabular}[c]{@{}l@{}}All other\\ moves that\\ are not \\ classified as \\ opening or \\ end game\\ moves\end{tabular}} & \begin{tabular}[c]{@{}l@{}}If total \# moves\textless{}=50 then end \\ game consist 35\% of  last moves \\ else 45\% of last moves \textbf{OR}\end{tabular} & \begin{tabular}[c]{@{}l@{}}Overall distribution of moves in \\ different phases and general \\ ideas\cite{van1982chess}\end{tabular} \\ \cmidrule(r){1-2} \cmidrule(l){4-5} 
\begin{tabular}[c]{@{}l@{}}\# of pieces exchanged\textless{}=8\\ \\ \textbf{OR}\end{tabular} & \begin{tabular}[c]{@{}l@{}}Initial exchanges have taken place and game \\ has moved to mid game \cite{chinchalkar1996upper}\end{tabular} &                                                                                                                                              & \begin{tabular}[c]{@{}l@{}}Less then 12 pieces remain\\ \\ \textbf{OR}\end{tabular}                                                                         & \begin{tabular}[c]{@{}l@{}}Board is simplified and both players \\ aim for strategic checkmate \\ \cite{dvoretsky2020dvoretsky,heinz1999endgame}\end{tabular}    \\ \cmidrule(r){1-2} \cmidrule(l){4-5} 
Both castling are available                                                         & \begin{tabular}[c]{@{}l@{}}If both players have done castling, game has\\  moved to mid game \cite{nimzowitsch2022my}\end{tabular}    &                                                                                                                                              & \begin{tabular}[c]{@{}l@{}}\# of legal moves for both \\ kings\textgreater{}=8 and both kings are in \\ third row (row 3 or 6)\end{tabular}        & \begin{tabular}[c]{@{}l@{}}King has taken a more active\\ role in the game \\ \cite{dvoretsky2020dvoretsky,heinz1999endgame}\end{tabular}                         \\ \bottomrule
\end{tabular}
}
\caption{Criteria used for categorizing chess moves into opening, midgame, or endgame phases. The rationale for each criterion is provided in separate columns for clarity.}
\label{tab_chess_game_segmentation}
\end{table*}

\section{Statistical test details}
\label{appendix_statistical_test}
As mentioned in Subsection \ref{subsec_statistical_test}, we have two text sources \textbf{(Sources, \(H\): Human, \(A\): AI)} and three segments from each text \textbf{(Segments, \(I\): Introduction, \(B\): Body, \(C\): Conclusion)}. \(Z_x\) is an individual feature extracted from segment \(x\) for source \(Z\).

For \textbf{source comparison} tests, we consider pairwise segments, \(x, y \in \{I, B, C\}\), compute their differences for human and AI texts, \(\Delta(H_x, H_y)\) and \(\Delta(A_x, A_y)\), respectively. Then, we address the key question, whether \(\Delta(H_x, H_y)\) differs significantly from \(\Delta(A_x, A_y)\) for any segment pair. We conduct a two-way ANOVA test ($\alpha = 0.05$) \cite{fisher1970statistical} focusing on the interaction effect of source (H vs. A) and cross-segment differences. If the interaction effect is significant, we proceed with post-hoc pairwise comparisons using the Wilcoxon signed-rank test. We opted for Wilcoxon signed-rank tests instead of t-tests due to the robustness to non-normal distributions \cite{hollander2013nonparametric}. These pairwise tests reveal whether human cross-segment differences \(\Delta(H_x, H_y)\) are statistically greater than (\(>\)), less than (\(<\)), or comparable (\(\sim\)) to AI cross-segment differences \(\Delta(A_x, A_y)\), for specific segment pairs. If no significant interaction effect is found in the ANOVA test, we infer that cross-segment differences between human and AI texts are not statistically meaningful.

Similarly, for  \textbf{segment comparison}, we compute the difference between human and AI texts for all three segments, \(\Delta(H_I, A_I)\), \(\Delta(H_B, A_B)\), and \(\Delta(H_C, A_C)\). Then, we conduct a one-way ANOVA test ($\alpha = 0.05$) with the three measures. If the result is statistically significant, we perform post-hoc pairwise comparisons between \(\Delta(H_x, A_x)\) and \(\Delta(H_y, A_y)\) for all segment pairs \(x, y \in \{I, B, C\}\). The post-hoc tests determine whether the human-AI feature difference is more pronounced in a specific segment or whether the differences are statistically indistinguishable across segments. If the ANOVA test shows no significant effects, we conclude that the differences between human and AI texts for the analyzed feature do not vary meaningfully across segments.

\section{Chess features extractions}
\label{appendix_chess_features}
Similar to segmenting text, dividing chess moves into opening, middlegame, and endgame can be subjective, as there are no strict rules for defining these transitions \cite{helfenstein2024checkmating}. While openings are identified by ECO codes, the middle game does not always begin immediately after these moves, nor can the start of the endgame be consistently determined by board conditions alone. Therefore, we draw on reasoning from existing studies \cite{horowitz1986win, van1982chess, chinchalkar1996upper, dvoretsky2020dvoretsky, heinz1999endgame, nimzowitsch2022my}, using factors such as piece counts, board conditions, and castling status to segment the games (Table \ref{tab_chess_game_segmentation}). To validate our rule-based method, we employ an LLM (\textit{GPT-4}) to segment a subset of $2000$ games, achieving a segmentation similarity score of $0.94$, indicating its effectiveness in approximating chess move segmentation.

\paragraph{Prompt for chess game segmentation}
\begin{center}
\small
    \fbox{%
    \begin{minipage}{\linewidth}
        \texttt{You are an expert in chess game understanding and moves. From the given list of moves you need to divide them into chess start, middle and end game moves. Your output should be strictly in the following format: \\
\{Start\}: <list of start game moves in comman seperated format>\\
\{Middle\}: <list of mid game moves in comman seperated format>\\
\{End\}: <list of mid game moves in comman seperated format>\\
moves list: \textbf{<original\_move\_list>}}

    \end{minipage}
}
\end{center}

Our next step involves creating a feature list from chess moves to computationally assess the differences between human and AI across game segments. While prior works have focused on cognitive aspects of chess play (e.g., memory, decision-making \cite{rasskin2009chess_metaphor}) or expert-driven analysis of key moments \cite{muller2018man_vs_machine}, recent advances in deep learning have enabled computational feature extraction in chess for tasks like next optimal move prediction, game outcome projection, and game clustering \cite{oshri2016predicting, brown2017machine, panchal2021chess}. Drawing on these studies, we extract 72 features related to board conditions, piece movements, positions, and captures. We also incorporate the optimal move and the corresponding player's win probability, as determined by the Stockfish engine \cite{stockfish} (with $time\_limit=0.1$ second) for each position.

\section{Text Features Extraction Details}
\label{appendix_text_features}

In this section, we discuss the details of extracting linguistic features from text that are essential to our analysis. For vocabulary richness, we consider the Brunét Index \cite{brunet1978vocabulaire}, as it is less sensitive to text length than the type-token ratio (TTR), making it more suitable for segments of varying lengths. For readability, we compute the Flesch Reading Ease score and employ the Python Textdescriptive library for additional linguistic insights.

\begin{figure*}[htbp]
    \centering
    \begin{subfigure}{0.3\textwidth}
        \centering
        \includegraphics[width=\linewidth]{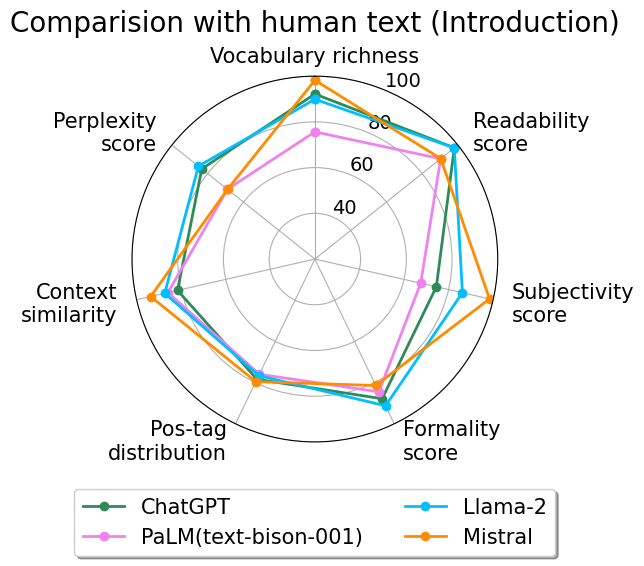}
    \end{subfigure}%
    \begin{subfigure}{0.3\textwidth}
        \centering
        \includegraphics[width=\linewidth]{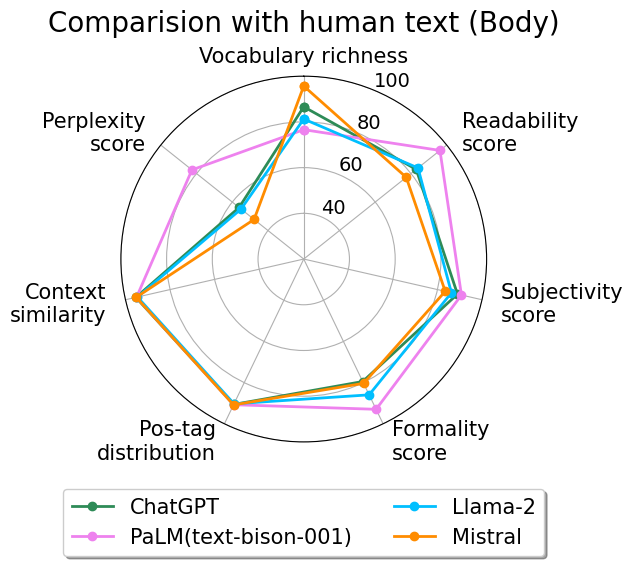}
    \end{subfigure}%
    \begin{subfigure}{0.3\textwidth}
        \centering
        \includegraphics[width=\linewidth]{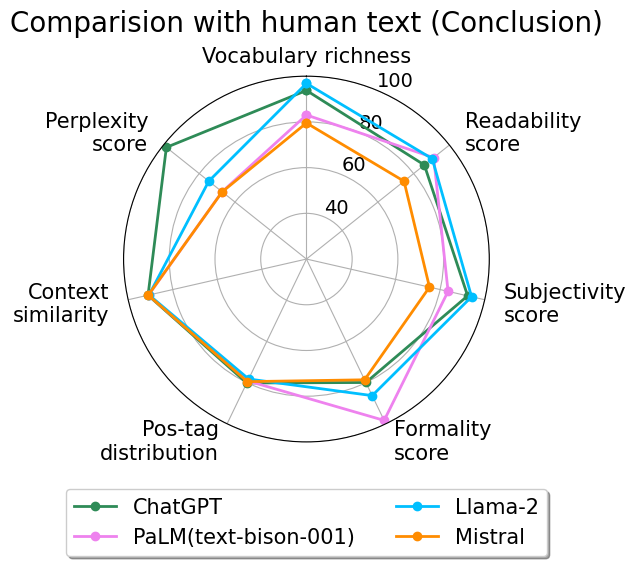}
    \end{subfigure}
    \caption{Comparison of individual LLMs to human text across segments. Values represent the similarity percentages (0-100) between AI text and human text for specific features, illustrating the extent to which individual LLMs can replicate human feature distributions.}
    \label{fig_segment_comparison_llm}
\end{figure*}

Syntactic features include part-of-speech (POS) tags, named entity recognition (NER), and stopword distributions extracted using SpaCy \cite{vasiliev2020spacy}. We further assess affective and stylistic elements through average sentiment and subjectivity scores using the VADER sentiment library, and formality scores via a pre-trained classifier \cite{formality_score_calculation}.

 For content analysis, we use OpenAI text embeddings (\textit{text-embedding-ada-002}) to capture the content within segments and measure the variation in embeddings between consecutive sentences or evaluate text predictability, we utilize GPT-2 to calculate both average perplexity and token-level perplexity scores, alongside burstiness, a metric that captures shifts in sentence structure and word choice. These features, shown to be impactful in recent AI text detection efforts \cite{GPTzero, venkatraman2023gptwho, mitchell2023detectgpt}, provide a comprehensive lens through which to explore the nuanced differences between human and AI-generated writing.

\section{Results for individual LLMs}
\label{appendix_individual_llm_result}

\begin{figure}[htbp]
    \centering
    \begin{subfigure}{0.25\textwidth}
        \centering
        \includegraphics[width=\linewidth]{Figures/fnr_rate_across_llm.png}
        \caption{}
    \end{subfigure}%
    \begin{subfigure}{0.25\textwidth}
        \centering
        \includegraphics[width=\linewidth]{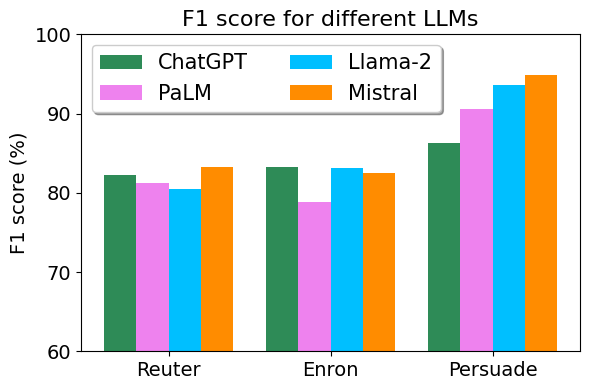}
        \caption{}
    \end{subfigure}
    \caption{\textbf{(a)} Analysis of False Negative Rates (FNR) across segments. Body shows lower FNR than introduction and conclusion for all LLMs. \textbf{(b)} F1 score comparison across datasets (using Finetuned BERT method). Essays demonstrate the highest F1 score, indicating more differences between human and AI-generated texts in this domain. Notably, F1 scores show minimal variation across different LLMs.}
    \label{fig_individual_llm_result}
\end{figure}

We also analyze how individual LLMs replicate human feature distributions across different text segments (Figure \ref{fig_segment_comparison_llm} and \ref{fig_individual_llm_result}). Overall, LLMs effectively mimic linguistic features, with the highest similarity observed in the body segments for most features, except for perplexity scores. ChatGPT demonstrates relatively balanced performance, while PaLM exhibits higher variability across segments. However, higher similarity scores do not necessarily imply that these LLMs are more challenging to detect, as detailed in the following subsection. Our analysis shows consistent performance across the three datasets. However, due to the shorter length of emails, they often lack clear structural distinctions. This results in some statistically insignificant findings in source and segment comparisons when contrasted with the other datasets.

\section{AI text detection methods}
\label{appendix_ai_text_detect}
\paragraph{GPTZero:} To determine whether a text is LLM-generated, GPTZero \cite{GPTzero} uses perplexity to measure the text's complexity and burstiness to evaluate sentence variants for providing the final output. We utilize the official API of GPT-Zero in our experiments.

\paragraph{MAGE:}MAGE (Machine-generated Text Detection in the Wild) is a Longformer model \cite{li-etal-2024-mage}, finetuned on the entire Deepfakedetect \cite{li2023deepfaketextdetect} dataset (comprising 447,674 human-written and AI texts). 
By effectively managing more than 512 tokens, Longformer \cite{Beltagy2020Longformer}, a modified Transformer architecture, gets around the drawbacks of conventional transformer models. Longer documents can be processed more easily because of their attention pattern, which scales linearly with sequence length. 
We also access the model from the HuggingFace repository\footnote{\url{https://huggingface.co/yaful/MAGE}
}.

\paragraph{RADAR:} RADAR is a robust AI text detection framework that leverages adversarial learning by jointly training a paraphraser and a detector \cite{hu2023radar}. The paraphraser aims to generate realistic, human-like text that can evade detection, while the detector learns to identify such paraphrased AI-generated content. In our study, we utilize the hosted version of RADAR available on Hugging Face\footnote{\url{https://huggingface.co/spaces/TrustSafeAI/RADAR-AI-Text-Detector}}.

\paragraph{Binocular:} Binoculars is a zero-shot, domain-agnostic method for AI text detection that operates without the need for training data \cite{hans2024binoculars}. It relies on cross-perplexity, computed as the cross-entropy between two language models that sharing the same tokenizer and vocabulary, when evaluated on a given text. Following the original implementation, we use the \textit{Falcon-7B} and \textit{Falcon-7B-Instruct} models for cross-perplexity computation in our experiments.

\paragraph{GPT-who:}GPT-who \cite{venkatraman2023gptwho} is a domain-agnostic statistical AI text detector that uses UID-based characteristics to capture unique statistical signatures. UID features are created via GPT2 inference and trained with a logistic regression model.
\paragraph{Finetuned-BERT:} We fine-tuned BERT (\textit{bert-base-cased}) on each dataset training set and evaluated it on the test set, as fine-tuned language models have been state-of-the-art in a lot of text classification and authorship tasks \cite{Tyo_Dhingra_Lipton_2022_AV}.

\section{AI text detection results for controlled settings}
As noted earlier, we conduct a length-controlled analysis to examine whether the middle portion of a text is more distinctive and contributes more to AI text detection. Tables \ref{table_ai_detection_C1} and \ref{table_ai_detection_C2} present the results for settings $C1$ (equal segmentation) and $C2$ (subsampled body matched to the length of the introduction and conclusion), respectively. In these experiments, detection is performed using only a specific segment of the text, along with voting across segment-level predictions. We exclude results using the total text, as they replicate the outcomes already reported for the original setting ($E$) in Table \ref{tab_ai_detection}.

\begin{table}[ht]
\centering
\resizebox{\columnwidth}{!}{
\begin{tabular}{llccccc}
\hline
\textbf{Dataset} & \textbf{Criteria} & \textbf{GPT Zero} & \textbf{MAGE} & \textbf{RADAR} & \textbf{Binoculars} & \textbf{GPT-Who} \\
\hline
\textbf{Reuters} & \textbf{Intro}       & 0.7309 & 0.7985 & 0.8186 & 0.7902 & 0.7201 \\
        & \textbf{Body}        & \textbf{0.7574} & \textbf{0.8271} & 0.8401 & 0.8576 & \textbf{0.7278} \\
        & \textbf{Conclusion}  & 0.7117 & 0.8263 & 0.8526 & 0.8572 & 0.7102 \\
        & \textbf{Voting}      & 0.7512 & 0.8026 & \textbf{0.8583} & \textbf{0.8965} & 0.7196 \\
\hline
\textbf{Enron}   & \textbf{Intro}       & 0.3772 & 0.8545 & 0.8290 & 0.1974 & 0.8127 \\
        & \textbf{Body}        & \textbf{0.6295} & 0.8193 & \textbf{0.8231} & \textbf{0.7052} & \textbf{0.8100} \\
        & \textbf{Conclusion}  & 0.5070 & 0.8612 & 0.8157 & 0.3552 & 0.8086 \\
        & \textbf{Voting}      & 0.4667 & \textbf{0.8760} & 0.8180 & 0.3237 & 0.8078 \\
\hline
\textbf{Persuade} & \textbf{Intro}      & 0.7972 & 0.7209 & 0.5021 & 0.7903 & 0.7862 \\
         & \textbf{Body}       & \textbf{0.8106} & \textbf{0.7423} & \textbf{0.5221} & 0.7864 & 0.7872 \\
         & \textbf{Conclusion} & 0.7320 & 0.7210 & 0.4908 & 0.8061 & 0.7857 \\
         & \textbf{Voting}     & 0.7990 & 0.7231 & 0.4821 & \textbf{0.8436} & \textbf{0.8336} \\
\hline
\end{tabular}
}
\caption{F1 scores of AI text detectors in the length-controlled setting ($C1$). Each value corresponds to the F1 score using the specified segment for a given dataset. \textbf{Bold} values highlight the segment or criterion achieving the highest F1 for each detector. Overall, the Body segment or Voting generally yields the best performance.}
\label{table_ai_detection_C1}
\end{table}

Across most datasets and detectors, the body (middle) segment consistently achieves higher F1 scores than the introduction or conclusion, reinforcing our findings from the original setting. It suggests that, even when length is controlled, the middle segment conveys stronger signals for distinguishing AI-generated text from human-written text. Moreover, voting shows improved performance compared to the original setting, highlighting its robustness. Finally, we note that results from $C1$ generally surpass those of $C2$ when using the body segment, since $C2$ preserves intact text segments rather than artificially truncated ones.

\begin{table}[ht]
\centering
\resizebox{\columnwidth}{!}{
\begin{tabular}{llccccc}
\hline
\textbf{Dataset} & \textbf{Criteria} & \textbf{GPT Zero} & \textbf{MAGE} & \textbf{RADAR} & \textbf{Binoculars} & \textbf{GPT-Who} \\
\hline
\textbf{Reuters}  & \textbf{Intro}      & 0.6167 & 0.7270 & 0.7222 & 0.7831 & 0.7186 \\
         & \textbf{Body}       & \textbf{0.7566} & \textbf{0.7475} & \textbf{0.7295} & 0.7608 & \textbf{0.7251} \\
         & \textbf{Conclusion} & 0.6943 & 0.7531 & 0.7292 & 0.8093 & 0.7237 \\
         & \textbf{Voting}     & 0.7158 & 0.7280 & 0.7213 & \textbf{0.8407} & 0.7186 \\
\hline
\textbf{Enron}    & \textbf{Intro}      & 0.3281 & 0.8660 & 0.8021 & 0.2021 & 0.8070 \\
         & \textbf{Body}       & 0.5962 & 0.7956 & 0.8030 & 0.3453 & \textbf{0.8076} \\
         & \textbf{Conclusion} & \textbf{0.6048} & 0.8580 & \textbf{0.8032} & \textbf{0.6489} & 0.8031 \\
         & \textbf{Voting}     & 0.4673 & \textbf{0.8713} & 0.8021 & 0.3066 & 0.8021 \\
\hline
\textbf{Persuade} & \textbf{Intro}      & 0.7092 & 0.5821 & 0.4410 & 0.7557 & 0.7604 \\
         & \textbf{Body}       & \textbf{0.8334} & \textbf{0.6512} & \textbf{0.4408} & 0.7571 & 0.7637 \\
         & \textbf{Conclusion} & 0.7809 & 0.6247 & 0.4398 & 0.8146 & 0.7756 \\
         & \textbf{Voting}     & 0.8085 & 0.5854 & 0.4398 & \textbf{0.8406} & \textbf{0.8142} \\
\hline
\end{tabular}
}
\caption{F1 scores of AI text detectors in the length-controlled setting ($C2$). We observe similar results like setting $C1$, Table \ref{table_ai_detection_C1}}
\label{table_ai_detection_C2}
\end{table}

\end{document}